\documentclass{article}

\usepackage{arxiv}

\usepackage[utf8]{inputenc} 
\usepackage[T1]{fontenc}    
\usepackage{hyperref}       
\usepackage{url}            
\usepackage{booktabs}       
\usepackage{amsfonts}       
\usepackage{nicefrac}       
\usepackage{microtype}      
\usepackage{lipsum}

\usepackage{floatrow}
\newfloatcommand{capbtabbox}{table}[][\linewidth]

\usepackage{graphicx}

\usepackage{float}
\usepackage{subfig}
\usepackage{algorithm2e}
\usepackage{amsmath}

\graphicspath{ {./images/} }

\title{Multi-Robot Path Planning Combining Heuristics and Multi-Agent Reinforcement Learning}

\author{
 Shaoming Peng \\
  School of Artificial Intelligence\\
  University of Chinese Academy of Sciences\\
  Beijing, China \\
  \texttt{pengshaoming2021@ia.ac.cn} \\
}

\begin{document}

\maketitle
\begin{abstract}
Multi-robot path finding in dynamic environments is a highly challenging classic problem. In the movement process, robots need to avoid collisions with other moving robots while minimizing their travel distance. Previous methods for this problem either continuously replan paths using heuristic search methods to avoid conflicts or choose appropriate collision avoidance strategies based on learning approaches. The former may result in long travel distances due to frequent replanning, while the latter may have low learning efficiency due to low sample exploration and utilization, and causing high training costs for the model. To address these issues, we propose a path planning method, MAPPOHR, which combines heuristic search, empirical rules, and multi-agent reinforcement learning. The method consists of two layers: a real-time planner based on the multi-agent reinforcement learning algorithm, MAPPO, which embeds empirical rules in the action output layer and reward functions, and a heuristic search planner used to create a global guiding path. During movement, the heuristic search planner replans new paths based on the instructions of the real-time planner. We tested our method in 10 different conflict scenarios. The experiments show that the planning performance of MAPPOHR is better than that of existing learning and heuristic methods. Due to the utilization of empirical knowledge and heuristic search, the learning efficiency of MAPPOHR is higher than that of existing learning methods.
\end{abstract}

\keywords{Multi-robot path planning \and multi-agent reinforcement learning \and heuristic search \and empirical knowledge}

\section{Introduction}
Multi-robot planning and vehicle navigation have wide applications in industrial, public, and residential settings, especially whenever the problem requires teamwork from robots. For instance, automated guided vehicle planning in workshops, mine automatic transportation vehicle navigation, parking lot vehicle path planning, and unmanned aerial vehicle transport path planning on airports are all common uses. These technologies can greatly improve efficiency and safety in various industries and settings.

Path planning involves two types of scenarios: planning in static environments and planning with obstacle avoidance strategies in dynamic environments \cite{van2007pathplanningdynamic}. Traditional path planning methods assume that obstacles remain static, making it difficult to solve problems with dynamic obstacles.  
The conventional approach to dealing with new obstacles is to use the D* or A* algorithm and replan the path \cite{cohen2017fastmap}. However, this strategy is time-consuming and fails to meet the real-time demands of most navigation scenarios. D* Lite \cite{koenig2002D_star_lite} is an algorithm that can efficiently optimize paths to cope with local changes, but it is not designed to handle multiple moving obstacles. As a result, D* Lite planning often fails when the environmental obstacles change significantly \cite{jin2023conflict_D_star_lite}.
In addition, traditional A* and D* algorithms do not take into account the shape and size of robots or their safe distances, which can often affect the success of path planning in practical applications.

In recent years, researchers have introduced reinforcement learning (RL) to solve dynamic multi-agent path-planning problems \cite{sartoretti2019primal, yang2020multi_agent_pp_dqn, ccetinkaya2021multi_agent_pp_drl}. Path planning algorithms based on RL can improve a robot's real-time responsiveness to dynamic obstacles, making it a promising method. Typically, RL-based path planning models are set up in simple grid environments and use a single agent for policy learning, which can then be extended to distributed multi-agent systems using single-agent models \cite{wang2020mobile_robot_pp_global_guided}. Although this approach is simplified, when faced with multiple moving agents at risk of collision, single-agent learning models may encounter non-stationary problems typically encountered by multi-agent reinforcement learning (MARL) methods.
In addition, optimizing path navigation strategies based on RL may encounter inherent problems such as sparse rewards and low sample utilization. 

To overcome the aforementioned challenges, we developed a path planning algorithm based on MARL, which combines a heuristic planner and a real-time MARL-based planner. Specifically, we first use a heuristic planner, D*, to obtain a globally optimal path called global guidance. During robot motion, the real-time planner tries to follow the global guidance as much as possible, detects dynamic obstacles around the global path, and gives actions based on local observations in real time to avoid collisions with other intelligent agents in motion.

The main contributions of this paper are as follows:
\begin{itemize}
    \item Firstly, a novel path-planning method that combines heuristic planning and real-time planning is proposed. The heuristic planner is responsible for motion planning, while the real-time planner is responsible for decision-making. During the robot's motion process, the real-time planner treats the heuristic planner as an action and decides whether to use it for re-planning based on its observations. To improve the efficiency of re-planning, the heuristic planner adopts the D* or A* algorithm.
    \item Secondly, a distributed real-time planning algorithm based on multi-agent reinforcement learning is proposed for dynamic path planning problems for multiple mobile robots. To address the non-stationarity of the environment caused by multiple mobile robots, a centralized training and distributed execution architecture is adopted. In the training stage, observation information is shared among the robots to assist in policy learning for each robot agent. In the execution stage, each robot calculates its action based on its own observations and action strategies. The MAPPO algorithm is applied to implement the above method in this paper.
\end{itemize}

\section{Related Work}

Dynamic multi-robot path planning algorithms can be classified into two categories: traditional heuristic planning methods and learning-based path planning methods. The former typically includes a heuristic function to assist with path optimization. 

\subsection{Heuristic Planning Approaches}
In response to dynamic environments, heuristic planning methods generally combine global planning and local planning, with the local planner responsible for modifying or optimizing the trajectory given by the global planner. For example, Brock et al. proposed a global dynamic window approach, which combines path planning and real-time obstacle avoidance, enabling robots to make reactive motion decisions in unknown and dynamic environments \cite{brock1999high_speed_dynamic_window}. However, this method may produce highly sub-optimal paths.
Mutib et al. proposed a multi-robot path planning algorithm based on D* Lite \cite{al2011D_star_lite_real_time_multi_agent}. In this algorithm, the obstacle map is dynamically updated by multiple collaborating robots, and the robot priorities are determined by the number of overlapping obstacles or minimizing the Euclidean distance. The performance of this algorithm decreases as the number of robots increases, but it is considered a feasible choice for a small group of robots.
Mehta et al. designed multiple navigation policies, including Go-Solo, Follow, and Stop, and used a joint-state-based cost function to determine the best policy, thereby achieving autonomous navigation in dynamic social environments \cite{mehta2016autonomous_dynamic_multi_policy}.
Xie et al. optimized aircraft path with the D* Lite algorithm, which utilized a priority selection strategy of two-level nodes to form a safe distance between obstacles and the planned path \cite{xie2020researchD_start_lite}.

Multi-robot path planning is essentially a Multi-Agent Path Finding (MAPF) problem, and the current mainstream algorithm for solving MAPF is the Conflict-Based Search (CBS) algorithm \cite{sharon2015conflict_CBS}. It is a leading two-level search algorithm for optimal MAPF solutions. Its core idea is to independently plan a path for each agent and resolve conflicts between two agents through branching. Researchers have made significant progress in accelerating CBS solving efficiency and expanding its applications \cite{boyarski2015ICBS, boyarski2020F_cardinal_CBS, barer2014suboptimal_ECBS, li2021EECBS, greshler2021cooperative_coCBS, jin2023conflict_D_star_lite}.

\subsection{Learning based Approaches}

Learning-based methods can achieve online real-time path planning, which is a promising approach to solve the problem of multi-mobile robot path planning. Among them, reinforcement learning is the mainstream method for solving robot path planning by completing tasks through trial and error. 
Yan et al. proposed an improved Double Deep Q Network (D3QN) method for online real-time path planning of unmanned aerial vehicles in dynamic environments \cite{yan2020towards_real_time_UAV}. Yang et al. utilized an improved Deep Q Network (DQN) algorithm to solve the problem of multi-AGV path navigation in warehouses \cite{yang2020multi_agv_path_planning}. Xu et al. proposed a path planning and dynamic collision avoidance algorithm for unmanned surface vehicles to follow COLREGs regulations in unknown marine environments \cite{xu2022pathCOLREGs}. Mert et al. trained a centralized DQN network to develop a multi-agent path planning strategy \cite{ccetinkaya2021multi_agent_pp_drl}. Chen et al. proposed a dynamic obstacle avoidance path planning method for robotic manipulators based on Soft Actor-Critic (SAC) algorithm in deep reinforcement learning \cite{chen2022deepdynamicobstacle}. Wang et al. proposed a distributed multi-agent path planning method that combines global guidance information with deep reinforcement learning and does not require agent communication \cite{wang2020mobile_robot_pp_global_guided}. These methods are mainly based on training a distributed planning algorithm using single-agent reinforcement learning. Shang et al. addressed the problem of collaborative scheduling of carrier-based aircraft and modeled the collaborative path planning of the aircraft group as a multi-agent reinforcement learning problem. They proposed a collaborative path planning method based on Multi-Agent Deterministic Deep Policy Gradient (MADDPG) algorithm \cite{shang2022collaborative_multi_carrier_based}.
However, applying reinforcement learning to multi-robot path planning tasks poses challenges due to complex states and sparse rewards. Therefore, researchers provide robots with intensive rewards through imitation learning and other methods to alleviate this problem \cite{sartoretti2019primal, li2020graph_neural_network, riviere2020glas_multi_robot_planning}. However, the quality of expert demonstrations and implied action policies may limit the robot's ability to learn better policies \cite{pmlr-v78-bhardwaj17a_learning_heuristic_search_imitation}.
Qiu used reinforcement learning to decide whether to execute traditional path-finding algorithms (such as A*) or other actions \cite{qiu2020multiMAPFRL}.

\section{Problem and Model}
\subsection{Environment}
The environmental map is modeled as a two-dimensional grid plane $\mathcal{M}\subseteq{\mathbb{R}^2}$ with the size of $H\times{W}$, which contains a set of $N^b$ static obstacles $\mathcal{B}=\{b_1, \cdots, b_{N^b}\}$, where $b_i$ denotes the $i{th}$ obstacle. The roadmap is represented by $\mathcal{F}=\left \langle \mathcal{V}, \mathcal{E} \right \rangle$, where $\mathcal{V}=\{v_1, \cdots, v_{N^f}\}= \mathcal{M}\setminus\mathcal{B}$ represents the set of $N^f$ free grids and $e_{ij}=(v_i, v_j)\in{\mathcal{E}}$ represents the minimum road segment between grid $v_i$ and $v_j$.
There are $N^a$ mobile robots $\mathcal{A}=\{a_1, \cdots, a_{N^a}\}$ in the environment. for each robot $a_i$, the other $N-1$ robots $\mathcal{A}^{-i}=\mathcal{A}\setminus\{a_i\}$ are dynamic obstacles to it. 
The space occupied by agent $a_i$ at time $t$ is represented by $C^i_t=\{s_1, s_2, \cdots \}\subset{\mathcal{V}}$, where $s_i$ denotes the grid occupied by agent $a_i$ at time $t$, then the grid space occupied by all other robots except for agent $a_i$ is represented by $C^{-i}_t={C^1_t}\cup\cdots\cup{C^{i-1}_t}\cup{C^{i+1}_t}\cup\cdots\cup{C^{N^a}_t}$. 
Therefore, the obstacles of agent $a_i$ at time $t$ are the sum of the static obstacles and dynamic obstacles, represented by $\mathcal{B}^{i}_t=\mathcal{B}\cup{C^{-i}_t}$, where any motion conflicts should be avoided. 

\subsection{Agent Model} \label{agentModel}
Robots are regarded as motion agents. Here, the robot is an abstract entity that can refer to automated guided vehicles in a warehouse, an airplane in an airport, a car in the parking lot, etc.
For agent modeling, we consider describing the grid space occupied by the agent. To reduce the computation during collision detection, we can utilize the physical characteristics of the agent and appropriately expand the space it occupies, establishing a convex polygon model of the agent \cite{si2021design_aircraft}. For instance, we can connect the convex points on the airplane outline to form a convex polygon (as shown in Fig. \ref{agent_model}), replacing the specific shape of the airplane with a convex polygon model. 
The shaded area of the convex polygon covering the grid map $C^i_t$, mentioned above, represents the grid space occupied by the agent. 

\subsection{Collision Detection Model} \label{collisionDetectionModel}

\begin{figure*}[!t]
\centering
\subfloat[Agent model.]{\includegraphics[width=0.5\textwidth]{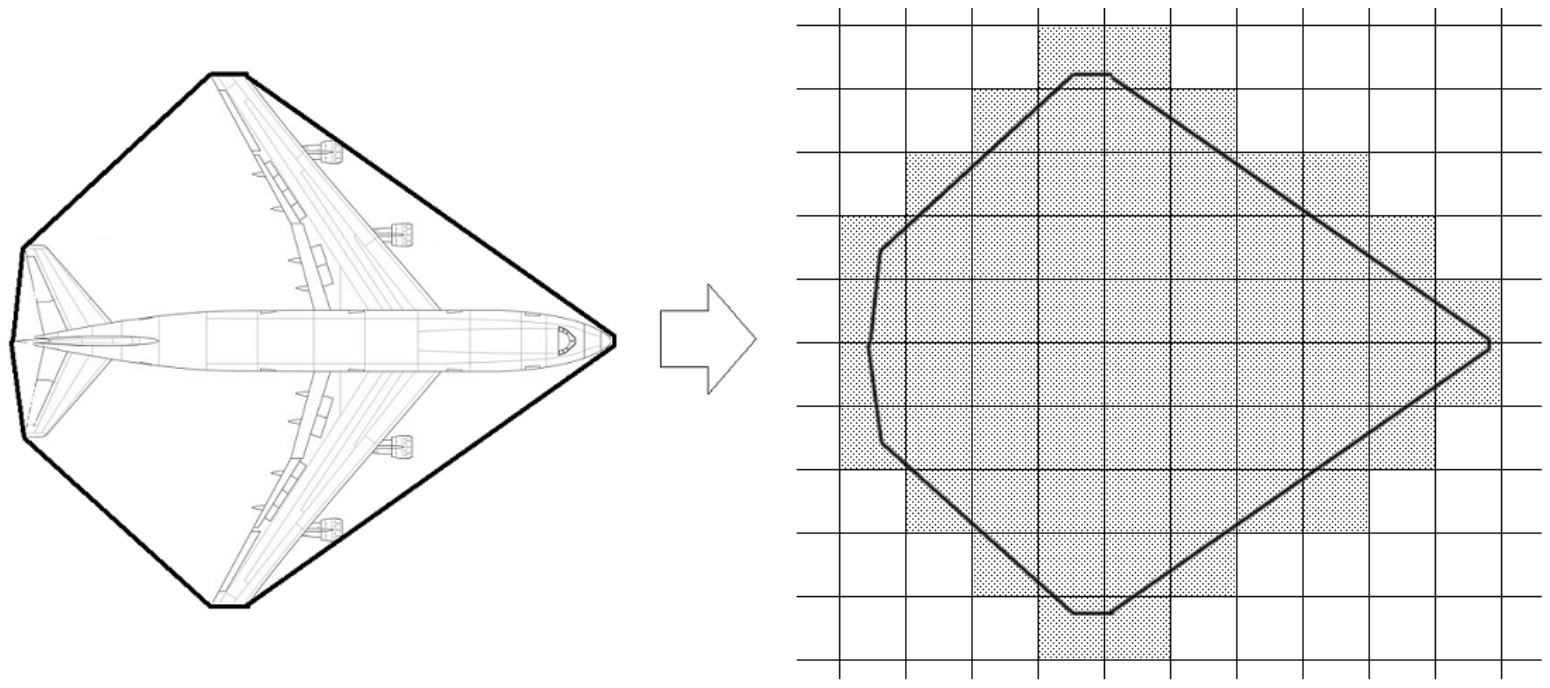}%
\label{agent_model}}
\hfil
\subfloat[Collision Detection Model.]{\includegraphics[width=0.5\textwidth]{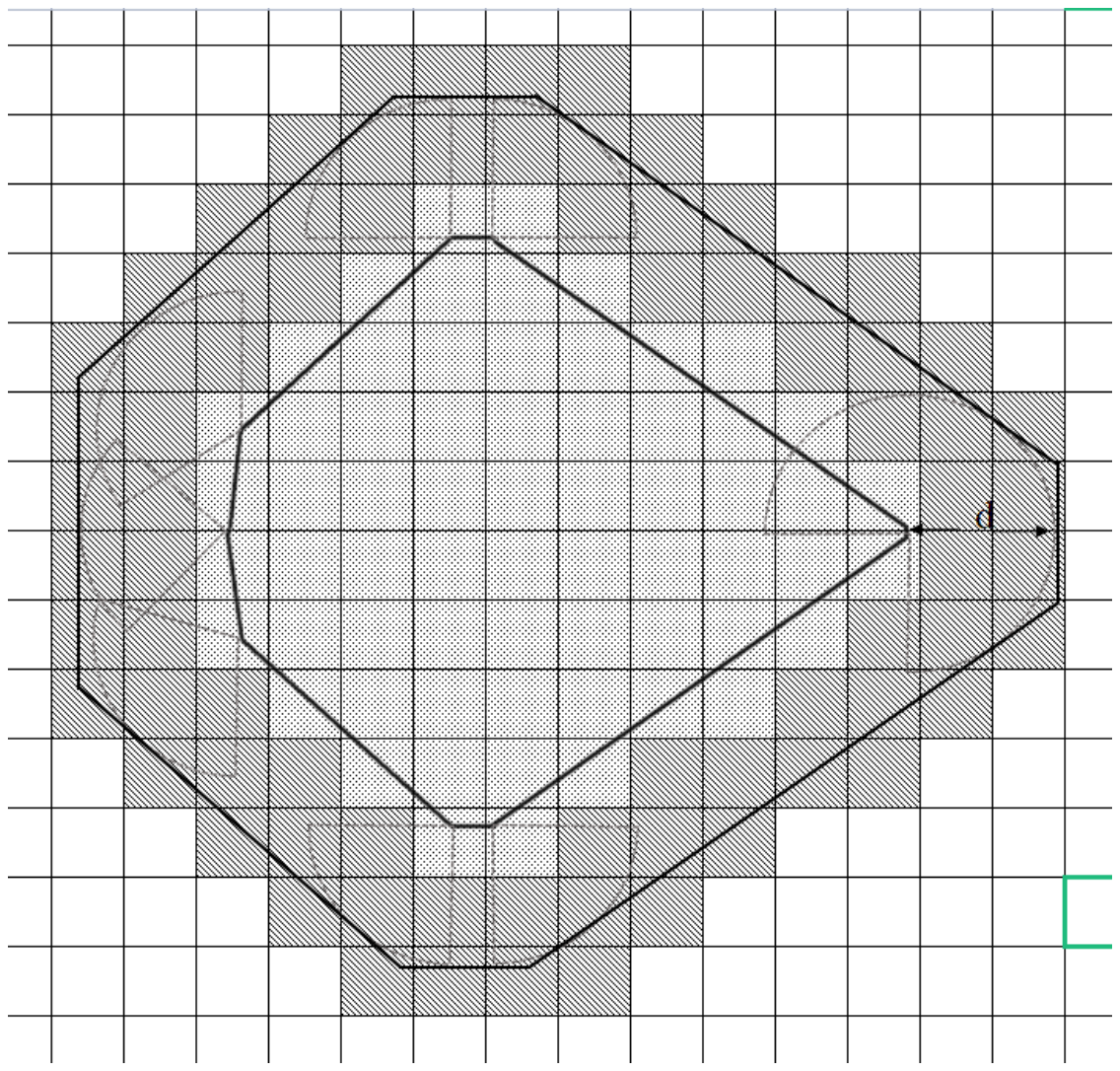}%
\label{safe_distance_area}}

\caption{Simulation comparison with D* Lite and MAPPOHR.}
\label{agent_and_collision_model}
\end{figure*}

To determine whether two agents collide, it is only necessary to calculate whether the shortest distance between the two agent models is equal to 0. According to Zhang et al.'s research \cite{zhang2014collision}, the convex polygon contour of the agent model is composed of a set of closed line segments. Therefore, the shortest distance between the line segment sets of two agent models is the shortest distance between the two agent models. However, for safety reasons, the encounter of two agents must consider ensuring a certain safe distance $d$ to ensure that there is no risk of collision. That is, if the shortest distance between two agent models is less than the safety distance $d$, it is considered a collision.
Correspondingly, a set of circles with a radius of the safety distance $d$ is drawn around the vertices of the convex polygon. The convex polygon formed by the tangents of all the circles is the collision detection model that considers the safe distance (see Fig. \ref{safe_distance_area}).
We denote the collision detection model of agent $a_i$ at time $t$ ${EC}^i_t$.

\subsection{Guidance}
Guidance is the shortest motion path planned by a heuristic algorithm, such as D* or A*, given the start and the goal.
This path is defined by $\mathcal{P}=\{p_0, p_1, \cdots, p_E\}$, where $p_0=v_{start}$ denotes the start position, $p_E=v_{goal}$ denotes the goal position, $p_i\in{\mathcal{F}}$ denotes the intermediate position on the path, $(p_i, p_{i+1})\in{\mathcal{E}}$. 
The path may be updated at time $t$ when the real-time planner gives the heuristic algorithm the instruction for replanning. The updated path at time $t$ is represented by $\mathcal{P}_t$.

\subsection{Action}
The agent action set is defined by $\mathcal{U}=\{Move, Wait, Back, Replan\}$. 
In this set, $Move$ instructs the agent to move forward to the next position along the guiding path, $Wait$ allows the agent to stay in the current position, $Back$ tells the agent to move back to the previous position that it passed earlier, $Replan$ permits the agent to perform a replanning using a heuristic algorithm (such as A* or D*) to find a new path. 

\subsection{Observation}
The agent's decision-making is modeled as a partially observable Markov decision process. Each agent has a local field of view where it observes the environment. 
The agent's observation at time $t$ within the finite view with the size of $H\times{W}$ grid map is represented by a vector $O_t=(o^g_t, o^d_t, o^p_t, o^n_t)^{\top}$. In this vector, $o^g_t$ denotes the distance from the current position to the goal, $o^d_t$ denotes the distance between the current agent and other agents, $o^p_t$ denotes whether obstacles exist along the first $n$ steps of the path guidance, $o^n_t$ denotes whether obstacles exist on the up and down $n$ steps of the current position.

\subsection{Objective}
The objective is to minimize the total length of all paths $L$ that all agents travel. $L_i$ denotes the length of the path the agent $a_i$ takes, then $L=\sum (L_1, \cdots, L_{N^a})$.

\subsection{Reward}

No matter how complex the environment is, a guided path can provide dense rewards for robots, accelerating the convergence of navigation strategies. Two penalty functions are defined. The first is the punishment for not reaching the target position, denoted by $P_1=-1/L$, where $L$ is the total length of the left guide path. The punishment types include waiting, replanning, and relative distance between the current and target positions, according to the defined action set. The second is the penalty for collisions between robots, denoted by $P_2$. Both $P_1$ and $P_2$ are negative.
A reward function is defined for reaching the target position, denoted by $R_1$. 
Therefore, the total reward function of $R$ can be represented by Equation \ref{rewardfunction}.  
\begin{equation}
\label{rewardfunction}
    R(s,a) = \begin{cases}
        -1/L &  \text{Move one step,} \\
        -P_1 &   \text{Colliding with obstacles,}\\
        R_1 &  \text{Reach the goal.}  \\
    \end{cases}
\end{equation}

For collaborative multi-robot routing environments, the final reward function is the average of all the robots' reward functions.

\section{Method}
\subsection{Algorithm Structure}

\begin{figure*}[ht]
\centering 
\includegraphics[width=18cm,angle=0]{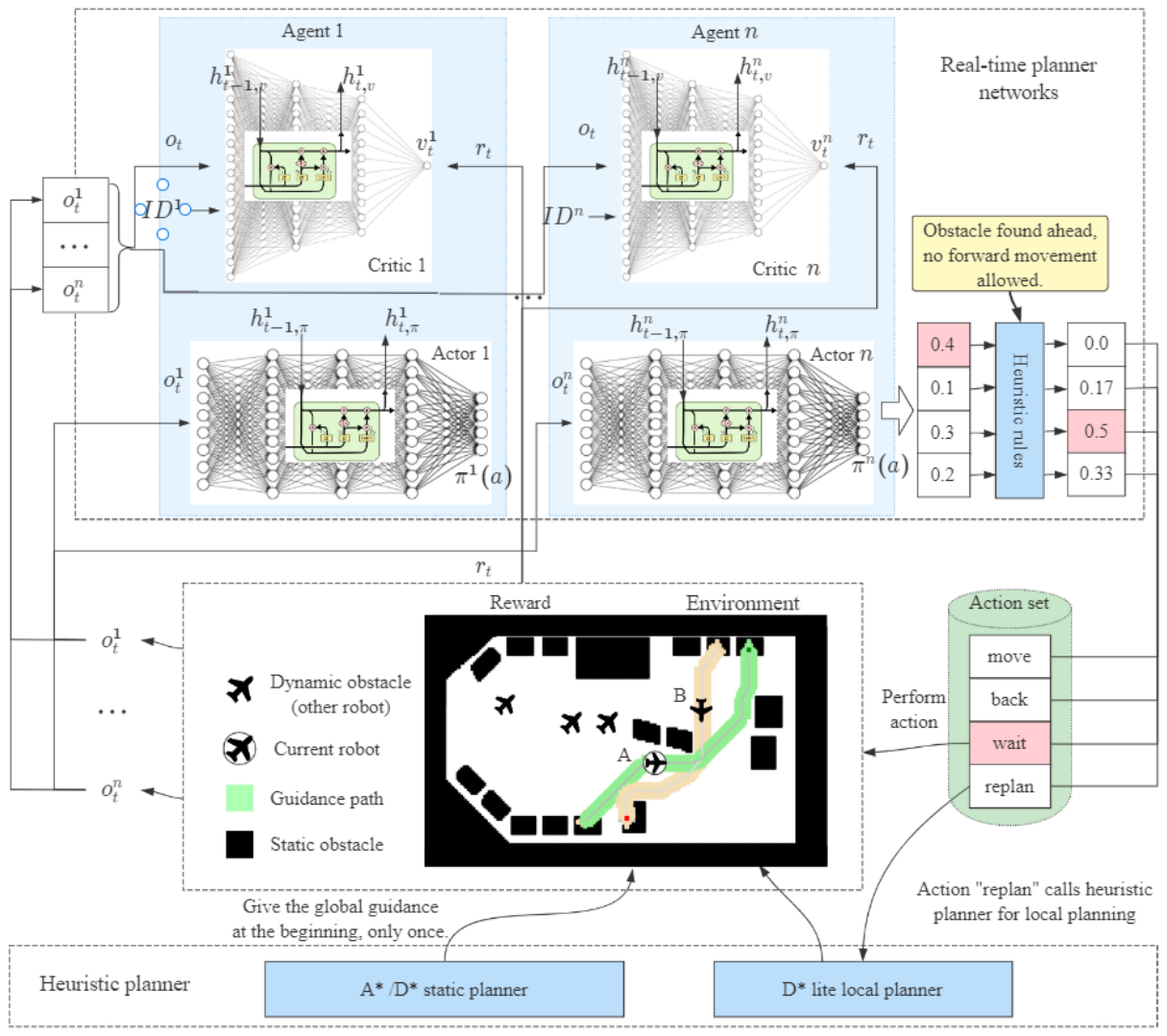}
\caption{Model structure.}  
\label{fig:systemStructure}
\end{figure*}

We propose a dynamic path planning algorithm that combines heuristic search and multi-agent reinforcement learning, called the MAPPOHR algorithm. The algorithm architecture consists of three components: environment model, heuristic search planner, and reinforcement learning-based real-time planner.

The environment model includes free space, static obstacles, and moving robots. When one robot moves, the other robots become its dynamic obstacles, and it should wait, retreat, or replan to find a new path to avoid collision with other robots. In Fig. \ref{fig:systemStructure}, the airplane represents the robot. The paths of robot A and robot B intersect at a point, and a collision is expected to occur when they both arrive at the intersection. Robot A may be able to adopt a waiting strategy to let robot B pass before proceeding.

The heuristic search planner consists of a static planner and a dynamic local planner. The static planner gives a global path for each robot, which does not consider dynamic obstacles and is only planned once at the beginning. The dynamic local planner provides local planning services for the robot during its movement. When the robot takes the ``replan" action, it calls the local planner to plan a new path to avoid colliding with other robots. The static planner can use the A* or D* algorithm, and the local planner adopts the D* Lite algorithm.

The real-time planner is based on multi-agent reinforcement learning. It requires a training process, and after training, it only needs to rely on the actor model to determine its own actions. The real-time planning model utilizes the MAPPO \cite{NEURIPS2022_9c1535a0_mappo} network structure, which includes $n$ agent networks, where $n$ denotes the number of robots. Each agent contains an actor network and a critic network, the former responsible for outputting actions and the latter responsible for evaluating the quality of actions. The actor network only inputs the agent's own observation information and outputs the probability distribution of actions. The input of the critic network shares the observation information of other agents to evaluate the agents' actions more scientifically. It should be noted that both the actor network and the critic network embed recurrent neural network layers to utilize historical information.

In addition, we added heuristic rules to the output layer of the real-time planner to avoid some actions inconsistent with common sense to output a more human-like probability distribution of actions.

\subsection{Heuristic Search}

\IncMargin{1em} 
    \begin{algorithm}
        \caption{procedure Cost($s,g$)} 
        \label{alg:DstarLiteCost} 
        
        Input: Starting point $s$, ending point $g$. \\
        Output: Cost from $s$ to $g$.
        
        Initialization: $s$ is the current position, and $g$ is the end position.
        
        \If {$ CollisionDetectionModel(s,g) \cap (staticObstacles \cup dynamicObstacles$)} 
            { return $\infty$. }
        \Else{return Distance($s,g$).}
        
    \end{algorithm}
\DecMargin{1em}

As mentioned above, both the global planner and the local planner in this article use heuristic search. The global planner and local planner in this article are both based on the D* Lite algorithm \cite{koenig2002D_star_lite}. The D* Lite algorithm describes a robot that occupies only one grid on the map and does not consider the safety distance between the robot and obstacles. The robot described in this article is an irregular shape that occupies multiple grids on the map. The agent model and collision detection model described in section \ref{collisionDetectionModel} are applied to describe the grid set occupied by the robot on the map and to calculate the cost function. The cost function is calculated as follows: if there is an intersection between the collision detection model of the robot and the static obstacles or dynamic obstacles, the robot is judged to have collided, and the cost function returns infinity. Otherwise, the cost is equal to the distance from the current position to the next position. The calculation method of the cost function is shown in Pseudocode \ref{alg:DstarLiteCost}, where the $CollisionDetectionModel$ function is used to calculate the collision grid set of the robot and the $Distance(s,g)$ function is utilized to calculate the distance from the current position $s$ to the next position $g$. The distance formula can use Euclidean distance or Manhattan distance, etc.
The heuristic search algorithm still follows the description in reference \cite{koenig2002D_star_lite}.

\subsection{Learning Algorithm}
We introduce MAPPO \cite{NEURIPS2022_9c1535a0_mappo} to implement the real-time planner. However, two points need to be mentioned:
(1) To improve the efficiency of training, the actor of each agent in MAPPO shares the network parameters.
(2) In addition to concatenating the observation information of other agents, the input vector of the critic network also adds specific features for the agent, including the agent ID and the relative distance from itself to other robots.

\subsection{Domain Knowledge}
We embedded heuristic rules in both the output layer of the learning model and the reward function. The heuristic rules used in the model output layer include the following:
\begin{itemize}
    \item When the agent reaches the endpoint or collides, its feasible action set is ``wait".
    \item When all other agents have reached the endpoint, and a robot's last planned path is longer than the global guide path, the robot replans the path and walks to the endpoint according to the new guide path. Otherwise, the robot walks directly to the endpoint according to the current guide path.
    \item Each robot explores $N$ steps forward according to its view of sight at every step. If there is no collision with other robots within $N$ steps, the robot moves forward according to the guide path. 
    \item If there is a collision risk in the next step along the guide path for a robot, the robot is forbidden to continue to move forward.
    \item If a robot's previous action is ``back" and there is still a collision risk in the next step, the robot moves forward.
    \item If all other robots wait, the current robot can move forward or replan.
\end{itemize}

The reward function embeds the following rule: A negative reward is given when the following conditions are satisfied,  
\begin{itemize}
    \item All robots execute ``wait".
    \item All robots execute ``replan".
    \item There is a collision risk between the two robots; one executes ``move" or ``back", and the other executes ``replan".
\end{itemize}

We call these heuristic rules domain knowledge or experience. 
These domain experiences are very helpful in improving training efficiency.

\section{Implementation}
\subsection{Environments}
We set up a path planning test scene in an airport. The airport is a rectangular ground with a size of $111\times171$, and there are static obstacles around and in the middle of the field, as shown in Fig. \ref{fig:systemStructure}. This test case sets two planes moving simultaneously, each with its own starting and ending positions. We set up 20 training cases and 10 test cases. Each test case contains two planes with different starting and ending points. The collision situation in each case is also different, with a collision rate of 0.9.

\subsection{Model Parameters}
In this experiment, the D* Lite algorithm generates the global guide path. A total of 30 global guide paths are generated for 30 cases. The size of the plane is $7\times7$, occupying 23 grids in the map. For the reward function, $P_2=-100$, and $R_1=200$. The discount rate of the reward function is 0.99. The learning model has two MLP layers and one RNN layer. The input feature dimension is 71, and the hidden layer dimension is 128. The activation function is ReLU. The clipping parameter of PPO is 0.1.
The re-planning time is set within 3 seconds.

\subsection{Training and Testing}
We train our model with one NVIDIA GeForce PTX 2070 GPU in Python 3.9 with torch 1.11.0+cu113. The learning rate is 1e-5, and the training optimizer is RMSprop. The total number of training steps is 1e5. In each episode, the maximum decision length of the robot is 60, and the robot will automatically stop if the decision length exceeds 60 steps. Last, the well-trained models are tested in 10 test cases. 

\subsection{Comparative Algorithms}
To verify the effectiveness of the proposed MAPPOHR, we compared it with D* Lite \cite{koenig2002D_star_lite} and MADDPG \cite{shang2022collaborative_multi_carrier_based}.

In addition, to verify the effectiveness of embedding heuristics and rules in the proposed algorithm, we conducted three ablation experiments, and the corresponding algorithms were named MAPPOH, MAPPO, and PPOHR. MAPPOH represents the model without embedded rules, MAPPO represents the model without embedded heuristics and rules, and PPOHR represents an independent distributed model without agent communication.

\subsection{Performance Metrics}
We use three metrics to evaluate the algorithm performance:
\begin{itemize}
    \item Added Moving Cost: It equals the actual moving path length minus the global guide path length. Replanning and backward actions may increase the actual moving path. Added Moving Cost is abbreviated as Added.
    \item Planning Cost: It comes from the time cost caused by the robot's replanning, which can be converted into a moving cost, that is, $ C_p = T_p\times v$, where $C_p$ denotes the total planning time, and $v$ denotes the robot's moving speed. 
    \item Waiting Cost: It comes from the cost caused by waiting during the robot's movement. The waiting cost is calculated by the number of waiting nodes. Waiting once adds one path node.
    \item Success Rate: It denotes the success rate of the well-trained model in the test cases. 
\end{itemize}

\section{Results}

\subsection{Compared with D* Lite}
The test results of D* Lite for 10 test cases are shown in Table \ref{Dstarlitewithheuristics}, and the test results of MAPPOHR are shown in Table \ref{MAPPORwithDstarlite}. Comparing Tables \ref{Dstarlitewithheuristics} and \ref{MAPPORwithDstarlite}, it can be found that MAPPOHR has significantly lower added moving cost and planning cost than D* Lite but has increased waiting cost. This is because MAPPOHR considers that, in some scenarios, waiting is more cost-effective than replanning, which is consistent with human decision-making logic. Therefore, the total moving cost of MAPPO is lower than that of D* Lite.

From the success rate perspective, D* Lite successfully planned all 10 cases, while MAPPOHR failed to plan case 0. This is due to the time cost of replanning set in MAPPOHR.

\begin{figure*}[!t]
\centering
\subfloat[DstarLite3]{\includegraphics[width=0.25\textwidth]{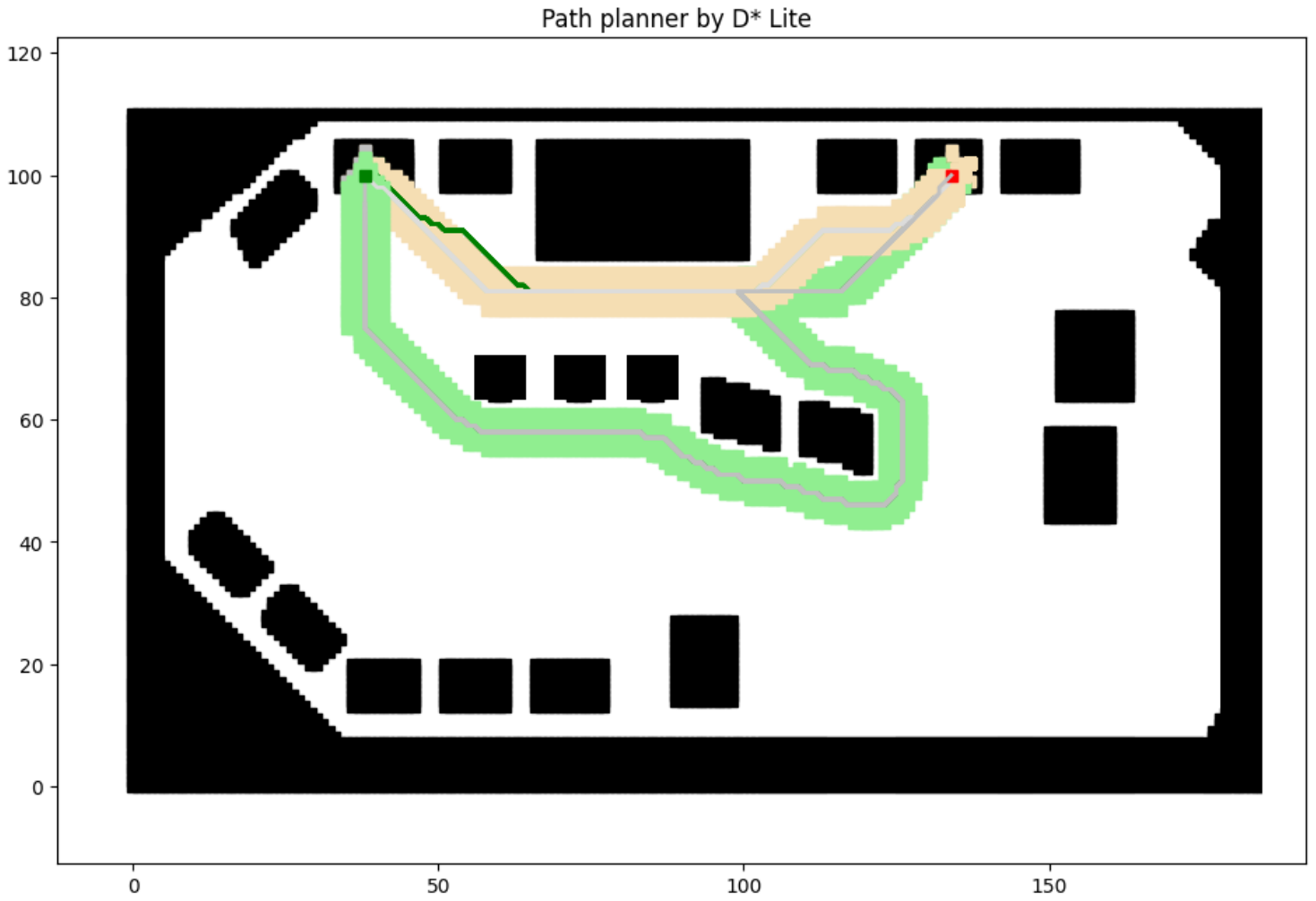}%
\label{DstarLite3}}
\hfil
\subfloat[MAPPOHR3]{\includegraphics[width=0.25\textwidth]{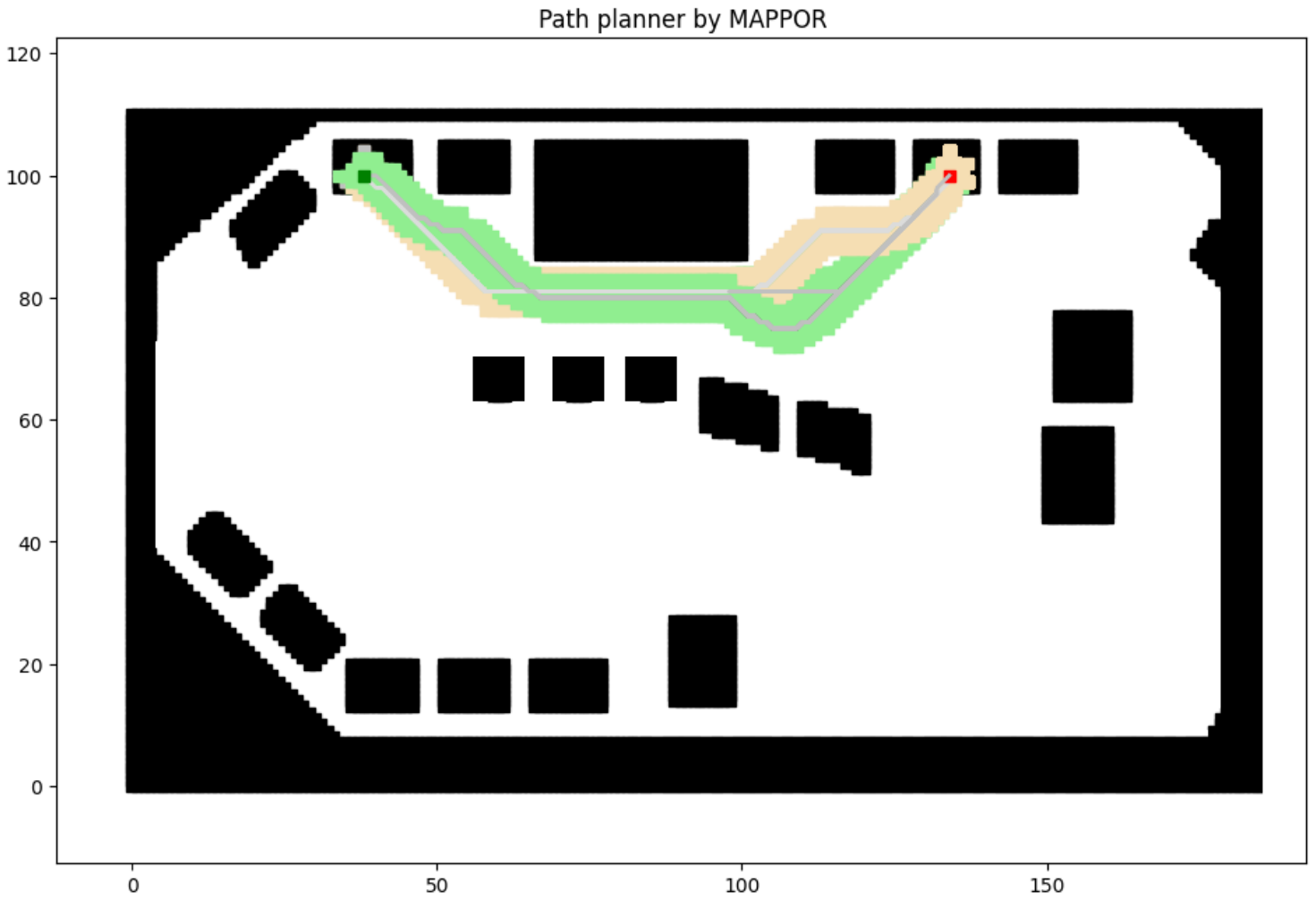}%
\label{MAPPOHR3}}
\hfil
\subfloat[DstarLite6]{\includegraphics[width=0.25\textwidth]{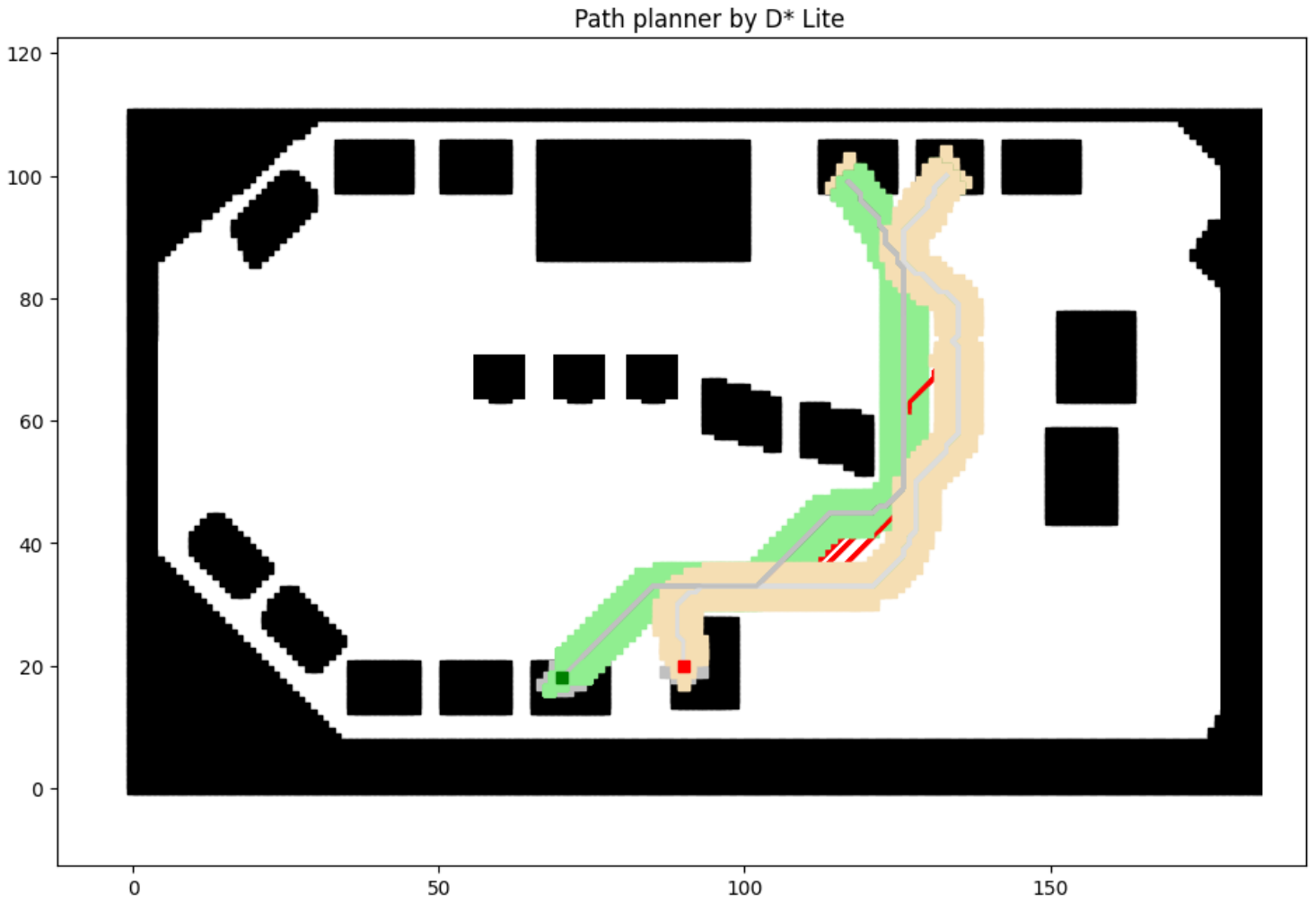}%
\label{DstarLite6}}
\hfil
\subfloat[MAPPOHR6]{\includegraphics[width=0.25\textwidth]{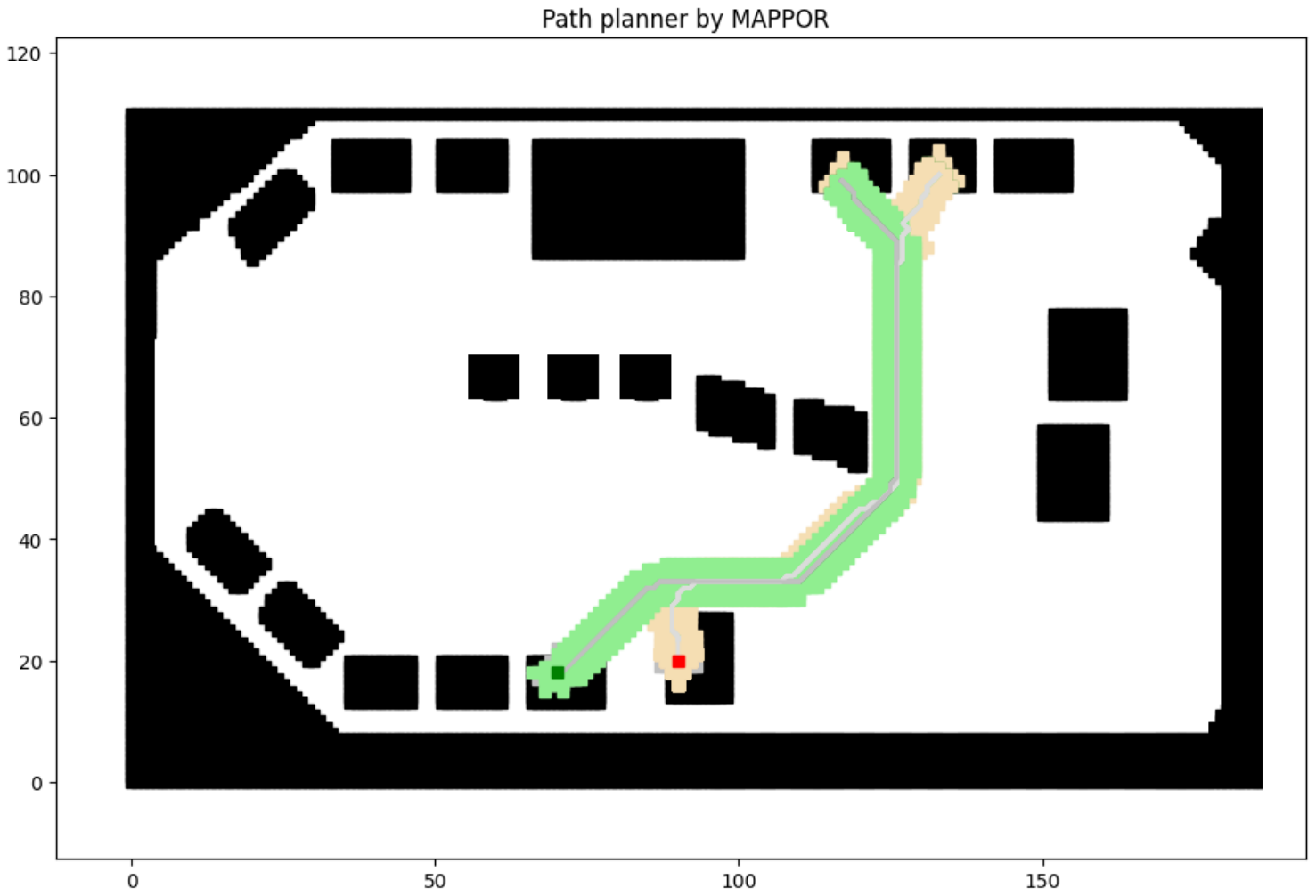}%
\label{MAPPOHR6}}
\hfil

\subfloat[DstarLite7]{\includegraphics[width=0.25\textwidth]{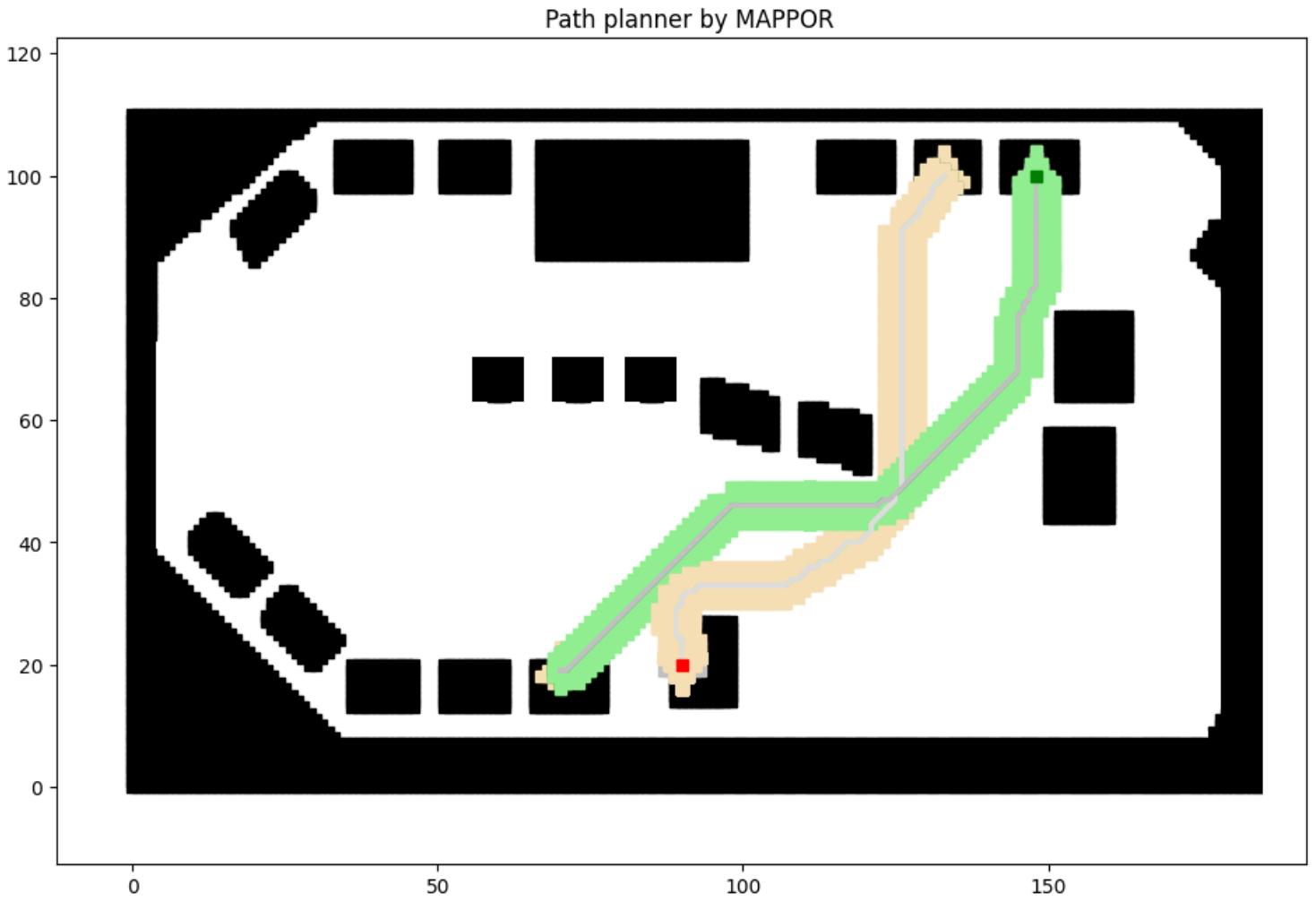}%
\label{DstarLite7}}
\hfil
\subfloat[MAPPOHR7]{\includegraphics[width=0.25\textwidth]{MAPPOR7.pdf}%
\label{MAPPOHR7}}
\hfil
\subfloat[DstarLite8]{\includegraphics[width=0.25\textwidth]{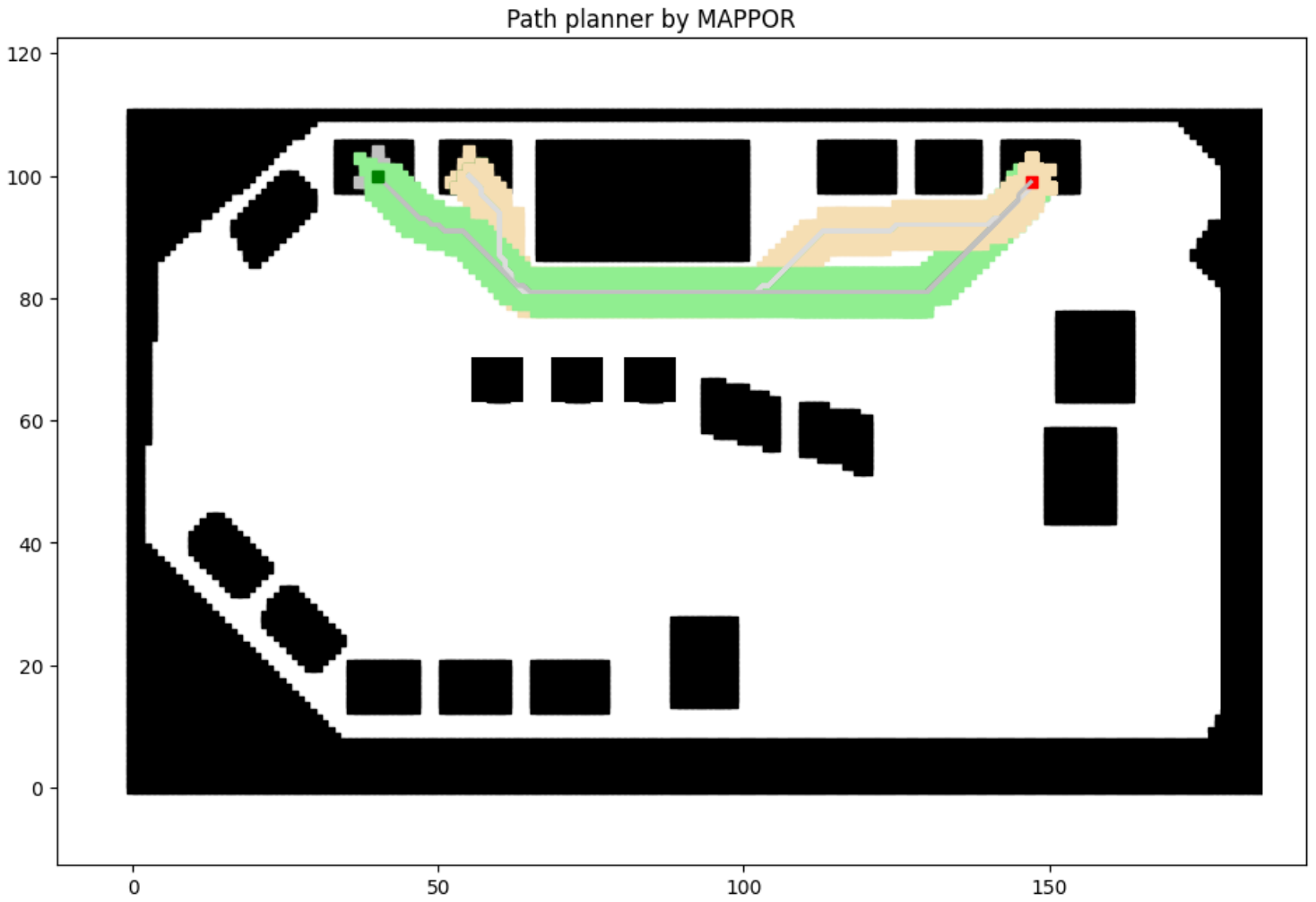}%
\label{DstarLite8}}
\hfil
\subfloat[MAPPOHR8]{\includegraphics[width=0.25\textwidth]{MAPPOR8.pdf}%
\label{MAPPOHR8}}
\hfil

\subfloat[DstarLite9]{\includegraphics[width=0.25\textwidth]{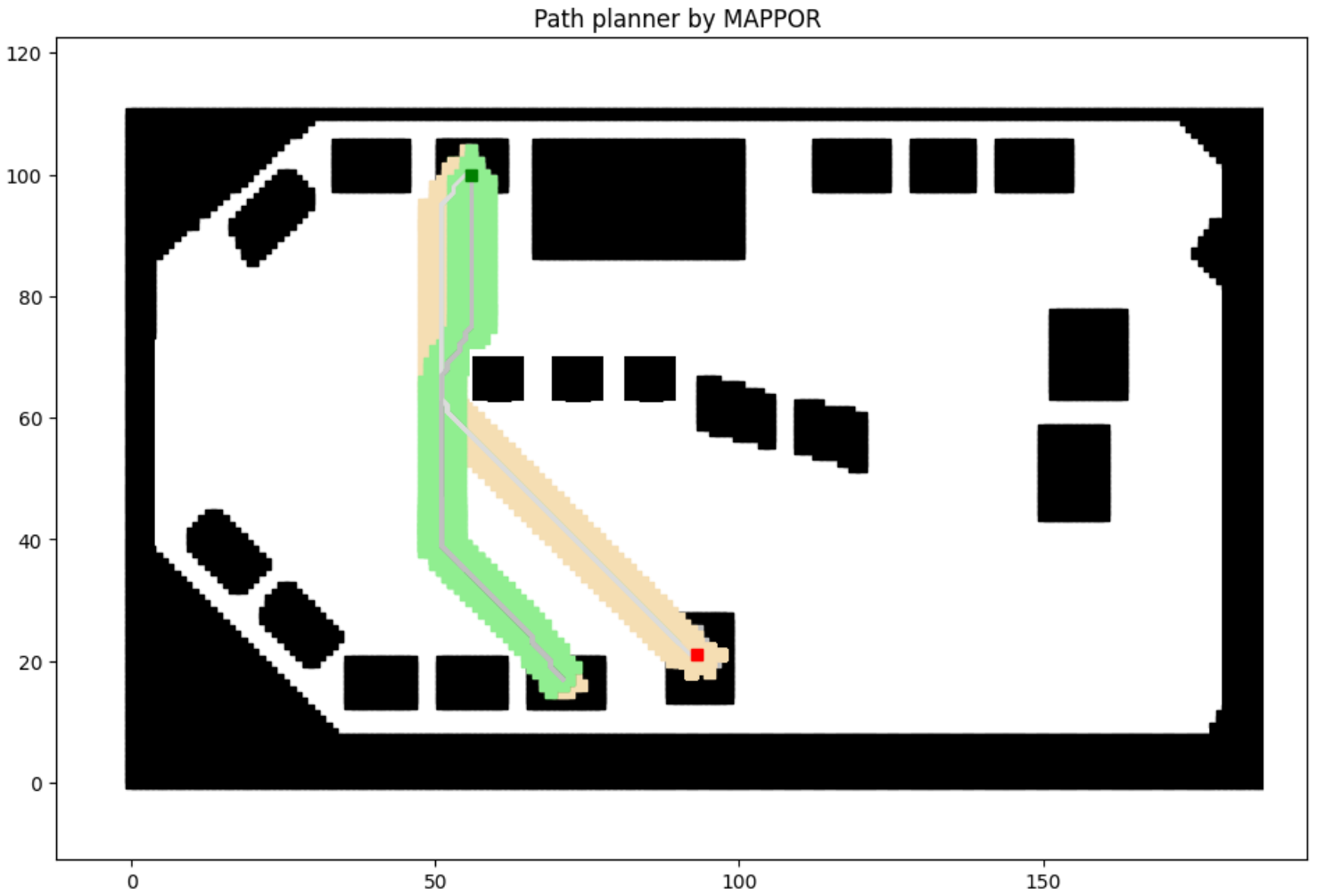}%
\label{DstarLite9}}
\hfil
\subfloat[MAPPOHR9]{\includegraphics[width=0.25\textwidth]{MAPPOR9.pdf}%
\label{MAPPOHR9}}

\caption{Simulation comparison with D* Lite and MAPPOHR.}
\label{fig_sim_Dstar_MAPPOHR}
\end{figure*}

We selected 5 test cases to show the actual planned moving paths of D* Lite and MAPPOHR, as shown in Fig. \ref{fig_sim_Dstar_MAPPOHR}. It can be seen that D* Lite does not consider other obstacle avoidance strategies except for replanning, so it frequently uses the replanning strategy, resulting in high replanning cost and long actual moving path, as shown in Fig. \ref{DstarLite3} and Fig. \ref{DstarLite7}. Correspondingly, MAPPOHR can choose more sensible strategies, such as giving way and waiting to avoid collisions with other robots based on on-site observations, thus saving moving costs.

\begin{table*}
\begin{floatrow}
\capbtabbox{
\begin{tabular}{c c c c c c}
\toprule
Case  & Added & Planning & Waiting & Total & Success \\
\midrule
0  &  28.48  &  14.77  &  0  &  43.25  &  1 \\
1  &  0  &  0.52  &  0  &  0.52  &  1 \\
2  &  0  &  0  &  0  &  0&  1 \\
3  &  105.84  &  81.01  &  0  &  186.85  &  1 \\
4  &  2.83  &  4.76  &  0  &  7.58  &  1 \\
5  &  0  &  0  &  0  &  0&  1 \\
6  &  25.9  &  66.33  &  0  &  92.23  &  1 \\
7  &  53.76  &  71.95  &  0  &  125.7  &  1 \\
8  &  14.04  &  17.52  &  0  &  31.56  &  1 \\
9  &  7.46  &  13.5  &  0  &  20.95  &  1 \\
\hline
Total  &  238.3   &  270.4   &  0.0   &  508.6  &  1.0  \\
Mean  &  23.8   &  27.0   &  0.0   &  50.9  &  - \\
\bottomrule
\end{tabular}
}{
 \caption{D* Lite test performance.}
 \label{Dstarlitewithheuristics}
}
\capbtabbox{
\begin{tabular}{cccccc}
\toprule
Case  & Added & Planning & Waiting & Total & Success \\
\midrule
0  &  0  &  0  &  61  &  61  &  0 \\
1  &  4  &  0  &  2  &  6  &  1 \\
2  &  0  &  0  &  0  &  0  &  1 \\
3  &  80.57  &  54.7  &  17  &  152.27  &  1 \\
4  &  2  &  0  &  11  &  13  &  1 \\
5  &  0  &  0  &  0  &  0  &  1 \\
6  &  0.83  &  0.51  &  27  &  28.34  &  1 \\
7  &  1.17  &  3.8  &  24  &  28.97  &  1 \\
8  &  48  &  0  &  3  &  51  &  1 \\
9  &  32.14  &  0  &  0  &  32.14  &  1 \\
\hline
Total  &  168.7   &  59.0   &  145.0   &  372.7   &  0.9 \\
Mean  &  16.9   &  5.9   &  14.5   &  37.3   &  - \\
\bottomrule
\end{tabular}
}{
 \caption{MAPPOHR test performance.}
 \label{MAPPORwithDstarlite}
 \small
}
\end{floatrow}
\end{table*}

\subsection{Compared with other MARL algorithms}
\begin{figure*}[ht]
\centering 
\includegraphics[width=0.99\linewidth]{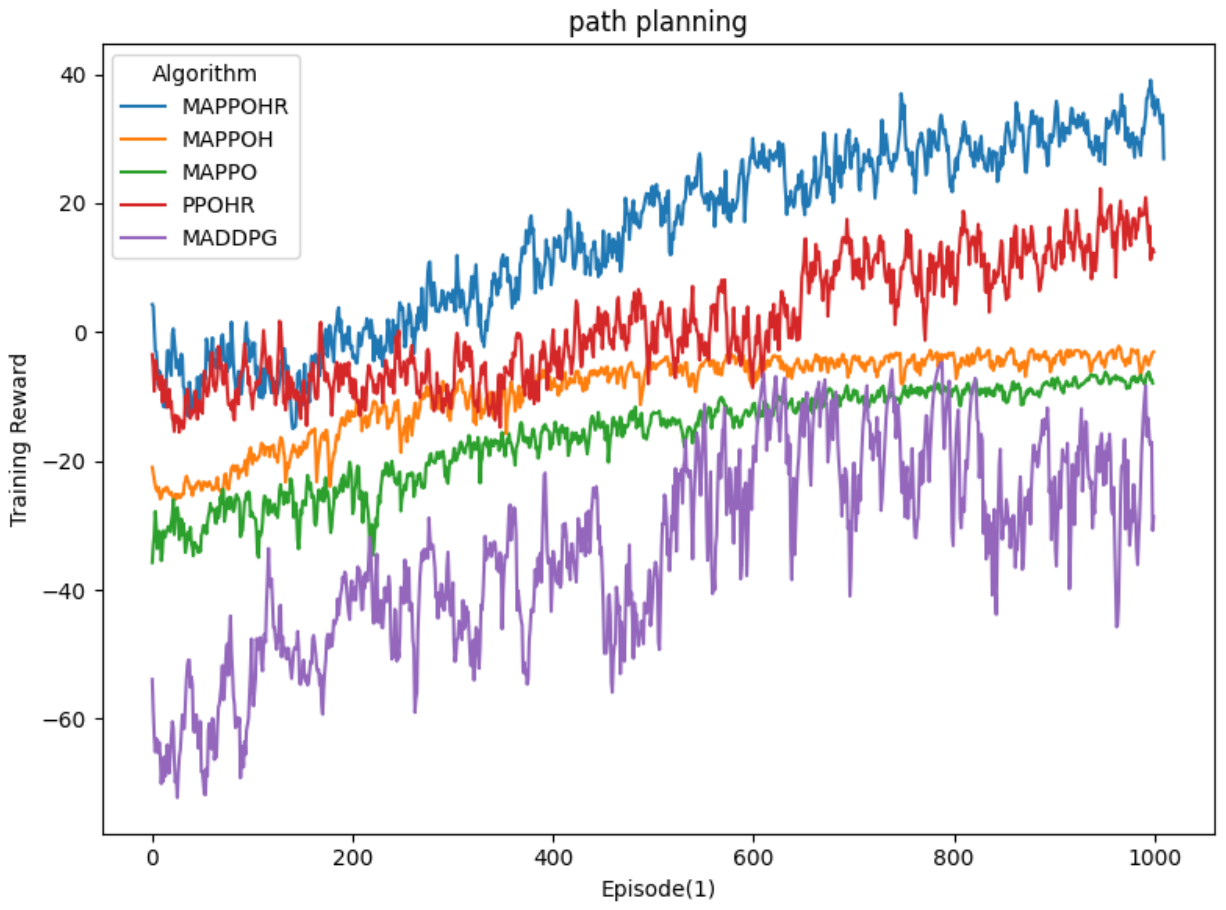}
\caption{MAPPOHR, compared with PPO, MAPPO, MADDPG.}  
\label{fig:rl_algos_comparation}
\end{figure*}

The training reward curves of MAPPOHR, MADDPG, MAPPO, MAPPOH, and PPOHR are shown in Fig. \ref{fig:rl_algos_comparation}.

Firstly, our method was compared with MADDPG. It can be seen from Fig. \ref{fig:rl_algos_comparation} that MAPPOHR obtained significantly higher rewards than MADDPG during the training process. According to the description of the MADDPG method in the literature \cite{shang2022collaborative_multi_carrier_based}, MADDPG adopts actions in eight directions such as up, down, left, and right, and the exploration space is very large. However, MAPPOHR only has four actions and combines global path guidance and heuristics. Its exploration space is relatively smaller and is conducive to learning action policy with prior knowledge. Therefore, the learning efficiency and effectiveness of MAPPOHR should be better than MADDPG.


Comparing the reward curves of MAPPOHR and PPOHR, we can see that the overall reward of MAPPOHR is higher than that of PPOHR. This indicates that sharing information with other agents is beneficial for learning action strategies.

Comparing the reward curves of MAPPOHR and MAPPO, we can find that the reward curve of MAPPOHR is much higher than that of MAPPO, indicating that integrating heuristics and prior rule knowledge in MAPPOHR is very helpful for learning.

Comparing the reward curves of MAPPOHR and MAPPOH, we know that the reward curve of MAPPOHR is also higher than that of MAPPOH, indicating that the utilization of prior rules is very beneficial for learning planning strategies.

\section{Conclusion and Future Work}
In this paper, we propose a path planning method, MAPPOHR, which combines heuristic search and multi-agent reinforcement learning for the multi-robot path-finding problem. The method consists of two layers: the upper layer is a real-time planner based on the multi-agent reinforcement learning algorithm MAPPO, while the lower layer is a heuristic search planner that guides the planning path. During initialization, it provides global path guidance, and during real-time navigation, it provides local path planning based on the instructions of the real-time planner. The experiments show that the planning performance of the MAPPOHR algorithm is better than that of pure learning methods based on MAPPO and MADDPG. The utilization of heuristics and empirical rules in the algorithm is beneficial for improving learning efficiency and effectiveness.

\bibliographystyle{unsrt}  
\bibliography{references}  






\end{document}